\documentclass[runningheads]{llncs}
\usepackage[T1]{fontenc}
\usepackage{graphicx}
\usepackage{multirow}
\usepackage{tabularray}
\usepackage{xcolor}

\usepackage{times}
\usepackage{amsmath}
\usepackage{algorithm}
\usepackage{algorithmic}
\usepackage[switch]{lineno}

\begin{document}
\title{ProtoGMM: Multi-prototype Gaussian-Mixture-based Domain Adaptation Model for Semantic Segmentation}
\titlerunning{ProtoGMM: Multi-prototype GMM-based Domain Adaptation Model}
%
\author{Nazanin Moradinasab\inst{1}\orcidID{0000-0003-3881-8599} \and
Laura S. Shankman\inst{2}\orcidID{0000-0003-4983-9467} \and
Rebecca A. Deaton\inst{2}\orcidID{0000-0001-7569-1924} \and
Gary K. Owens\inst{3}\orcidID{0000-0002-7119-9657}\and
Donald E. Brown\inst{1,4}\orcidID{0000-0002-9140-2632}}
\authorrunning{N. Moradinasab et al.}
%
\institute{Department of Engineering Systems and Environment, University of Virginia, Charlottesville, VA, USA \and
Laboratory of Dr. Gary Owens Cardiovascular Research Center, University of Virginia, Charlottesville, VA, USA \and
Department of Molecular Physiology and Biological Physics, University of Virginia, Charlottesville, VA, USA
\and
School of Data Science, University of Virginia, Charlottesville, VA, USA\\
\email{\{nm4wu,lss4f,rad5x,gko,deb\}@virginia.edu}}
\maketitle              
\begin{abstract}
Domain adaptive semantic segmentation aims to generate accurate and dense predictions for an unlabeled target domain by leveraging a supervised model trained on a labeled source domain. The prevalent self-training approach involves retraining the dense discriminative classifier of $p(class|pixel feature)$ using the pseudo-labels from the target domain. While many methods focus on mitigating the issue of noisy pseudo-labels, they often overlook the underlying data distribution p(pixel feature|class) in both the source and target domains. To address this limitation, we propose the multi-prototype Gaussian-Mixture-based (ProtoGMM) model, which incorporates the GMM into contrastive losses to perform guided contrastive learning. Contrastive losses are commonly executed in the literature using memory banks, which can lead to class biases due to underrepresented classes. Furthermore, memory banks often have fixed capacities, potentially restricting the model's ability to capture diverse representations of the target/source domains. An alternative approach is to use global class prototypes (i.e. averaged features per category). However, the global prototypes are based on the unimodal distribution assumption per class, disregarding within-class variation. To address these challenges, we propose the ProtoGMM model. This novel approach involves estimating the underlying multi-prototype source distribution by utilizing the GMM on the feature space of the source samples. The components of the GMM model act as representative prototypes. To achieve increased intra-class semantic similarity, decreased inter-class similarity, and domain alignment between the source and target domains, we employ multi-prototype contrastive learning between source distribution and target samples. The experiments show the effectiveness of our method on UDA benchmarks.

\end{abstract}
\section{Introduction}

In recent years, there has been remarkable progress in semantic segmentation, a technique that assigns semantic class labels to each pixel in an image. However, achieving the generalization of deep neural networks to unseen domains is vital for critical applications, including autonomous driving \cite{treml2016speeding}, and medical analysis \cite{wang2019automatic}. Unfortunately, this progress heavily depends on acquiring large-scale pixel-level annotations, a costly and time-consuming process when done manually.

\begin{figure}[h]
\centering
\includegraphics[width=.55\textwidth]{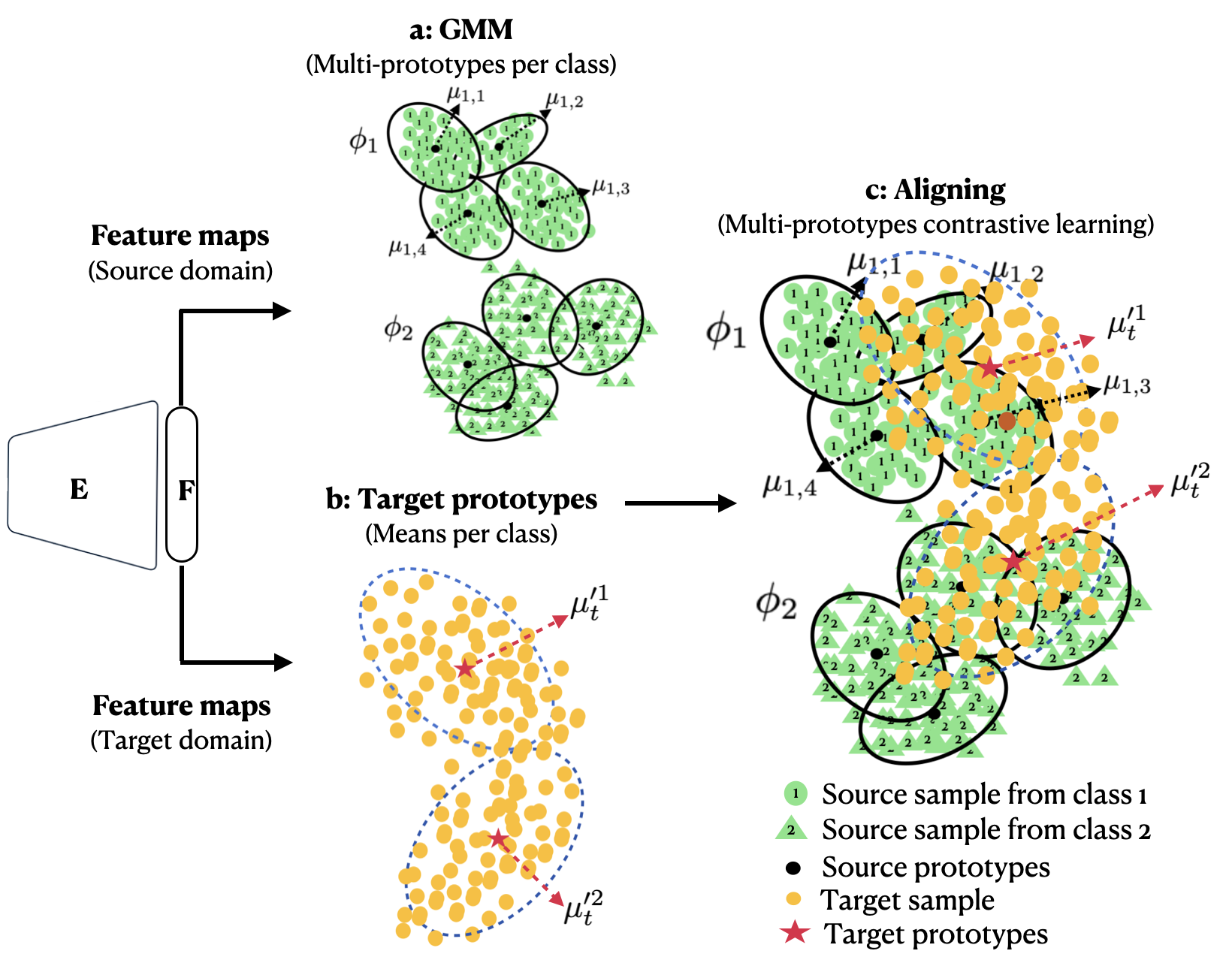} %
\caption{Diagram of Proposed Approach}
\label{fig:approach}
\end{figure}
To address this challenge, researchers have been exploring alternative approaches like generating simulated data or leveraging out-of-domain (source) annotations to reduce manual effort and improve neural network applicability across domains. However, domain shift remains a major obstacle, leading to a performance decline when applying a well-trained model from the source domain to the target domain. To tackle this issue, a solution known as unsupervised domain adaptation (UDA) has been proposed, which transfers knowledge from a label-rich source domain (synthetic) to a label-scarce target domain (real) \cite{wang2022cluster}. Recent trends in UDA for semantic segmentation have led to two main approaches: domain alignment and self-training.

Domain alignment methods use adversarial learning to address domain shift across spaces like image level \cite{li2019bidirectional}, feature level \cite{chen2017deeplab}, and output level \cite{luo2019taking}. While effective in global alignment, these methods may not ensure discriminative feature representations for various classes in the target domain.
Alternatively, self-training utilizes target-specific knowledge with high-confidence pseudo-labels in subsequent training rounds. Despite promising performance, these methods suffer from
significant limitations. Source domain bias results in unreliable pseudo-labels for the target domain, and relying solely on highly confident predictions provides limited supervision information during training. Effectively managing noisy labels and source domain bias is crucial. Some studies address this through techniques like confidence estimation \cite{corbiere2021confidence}, and label denoising \cite{zhang2021prototypical}.

In many papers, classifiers are often trained using cross-entropy loss, which emphasizes bringing similar features together but does not effectively differentiate features across distinct classes. It is essential to properly segregate features from distinct classes while aggregating those from the same class. In this context, Contrastive Learning (CL) emerges as a relevant topic, allowing models to learn meaningful visual representations by comparing diverse unlabeled data \cite{he2020momentum}. 
A simple CL method uses a memory bank, updated during training by adding averaged features of each category from the current source image and removing the oldest ones. This approach may be prone to class biases, updating underrepresented classes less frequently, and can be computationally expensive. 
Another approach is to use global prototypes, which are the averaged features of each category across the entire source domain. However, it overlooks variations in attributes (e.g., shape, color), reducing discriminability. Furthermore, it relies on the unimodality assumption of each category. While CL improves domain adaptation, it depends on pseudo-labels generated by a discriminative classifier trained with cross-entropy loss, causing issues with unreliable predictions in early training stages.
Vayyat \textit{et al.} (\cite{vayyat2022cluda}) address this by introducing a weight on the loss based on the confidence of the teacher-network predictions. However, the teacher network, a discriminative classifier, is biased toward source domain. Discriminative classifiers have limitations: neglecting the data distribution, assuming unimodality per class, and suffering accuracy degradation away from decision boundaries, hindering adaptation for critical tasks.

In this paper, to address these challenges we present a Multi-prototype Gaussian-Mixture-based model (ProtoGMM) that overcomes the limitations of existing methods. Unlike traditional self-training approaches focusing solely on the discriminative classifier of $p(class|pixel feature)$, ProtoGMM adopts a hybrid training approach, combining discriminative and generative models. The core of ProtoGMM lies in modeling the underlying distribution of source pixel features using generative Gaussian Mixture Models (GMMs) optimized through Expectation-Maximization (EM) (Figure \ref{fig:approach}-a). This approach enables effective adaptation to multimodal data densities. 
The GMM components serve as representative prototypes for CL losses. Rooted in the domain closeness assumption \cite{shi2012information}, ProtoGMM departs from using noisy pseudo-labels, determining positive and negative clusters for the target sample based on the underlying distribution of the source domain and target domain category prototypes.
By incorporating the GMM-based model with CL loss, ProtoGMM functions as a generative model, significantly improving the performance of the domain adaptation model when used alongside the discriminative classifier. In addition to its advantages in CL, ProtoGMM excels at addressing the label distribution shift, a common challenge in UDA tasks. 

\textbf{Contributions:} Our contribution involves introducing the ProtoGMM model for UDA in semantic segmentation. We address biases inherent in discriminative classifiers by combining a generative model with a discriminative model. Additionally, we showcase the superior performance of our approach compared to the current state-of-the-art (SOTA) approaches in two distinct scenarios: 1) synthetic-to-real adaptation, demonstrated through GTA $\rightarrow$ Cityscapes and Synthia $\rightarrow$ Cityscapes, and 2) cell-type adaptation in immunofluorescent images, where each cell type serves as an individual domain. In the latter scenario, we highlight the effectiveness of the ProtoGMM model in improving detection performance across different cell types.




\section{Background}

\textbf{Adversarial training: }
Various techniques address the distribution disparity between the source and target domains, targeting the pixel level, feature level, and output level through adversarial learning. 
Tsai \textit{et al.} (\cite{tsai2018learning}) discover that aligning the distribution of the output features yields greater effectiveness compared to aligning the distribution of the intermediate feature space. Nonetheless, the direct alignment of feature distributions in a high-dimensional space presents challenges. Sankaranarayanan \textit{et al.} (\cite{sankaranarayanan2018learning}) tackle this by reducing feature dimensions that contain the essential feature components. Long \textit{et al.} (\cite{long2018conditional}) propose that the global distribution alignment can compromise the distinctiveness of features within the target domain. To address this issue, Wang \textit{et al.} (\cite{wang2020classes}) incorporate the class information into the adversarial loss. However, adversarial training often encounters issues with stability because of the lack of a comprehensive understanding of each category. As a result, some studies opt for the utilization of category anchors \cite{wang2020differential} derived from source data to enhance the alignment process. However, 
selecting category anchors poses a challenge, and constructing global anchors based on the unimodal distribution assumption overlooks within-class variation. We suggest estimating the multi-prototype source distribution by applying GMM on the source feature space. GMM components then function as representative prototypes, accommodating multimodal data density and capturing within-class variations.
\noindent
\textbf{Self-training: }
In self-training approaches, pseudo labels are assigned to samples from the target domain to facilitate iterative training. The central concern in these techniques is how to achieve stable model training in the presence of noisy pseudo labels. Some studies proposed a variety of strategies such as dynamic threshold strategy \cite{zou2019confidence}, or uncertainty estimation strategy \cite{zheng2021rectifying}, to select high-quality pseudo labels. Recently, Zhang \textit{et al.} (\cite{zhang2021prototypical}) use the relative feature distance to the prototypes to refine further target pseudo labels. Most existing self-training approaches rely on training the classifier, i.e. discriminative model, using the cross-entropy loss function over the source ground truth and target pseudo labels. However, the ProtoGMM model incorporates the GMM generative model alongside the discriminative classifier. The ProtoGMM involves the generative branch in addition to the classifier. The generative branch is optimized using the CL loss guided using multi-prototype GMM. Our approach utilizes the knowledge of the underlying feature distributions alongside the classifier to boost the performance.

\section{Methodology}

\subsection{Problem formulation}
In the UDA, $p_{s}(x,y) \in p_{\mathcal{S}}$ and $p_{t}(x,y) \in p_{\mathcal{T}}$ are the underlying source and target domain distributions, respectively. Then, the labeled data $\mathcal{D}_{S}$ (i.e. $x_{s} \in R^{H \times W \times 3}$ and $y_{s} \in R^{H \times W \times C}$) is sampled i.i.d from the source domain distribution (i.e. $p_{s}(x,y)$) and unlabeled data $\mathcal{D}_{T}$ (i.e. $x_{t} \in R^{H \times W \times 3}$) is selected i.i.d from marginal target domain distribution (i.e. $p_{t}(x)$). Here, $H$ and $W$ represent the height and width of the images, respectively, and $C$ denotes the number of classes. The primary goal of UDA is to train the model using both $\mathcal{D}_{T}$ and $\mathcal{D}_{S}$ to enhance the model's performance on the target domain. The model itself is composed of three components: an encoder (E), a multi-class segmentation head (H), and an auxiliary projection head (F). When given an input image $x$, the auxiliary projection head processes the encoder's output to obtain a feature map ($f = F(E(x))$). All features are mapped to the $l2$-normalized feature vector. Subsequently, the multi-class segmentation head operates on the encoder's output to produce a class probability map ($\hat{y} = H(E(x))$). 
We utilize the cross-entropy loss ($L_{ce}$) and ProtoGMM loss functions to train the model. 
The cross-entropy loss is computed for the source and target domain images using their ground truth labels ($y_{i}^{s}$) and pseudo labels ($\hat{y}^{t}_{i}=\underset{c}{argmax}\:\:\hat{y}_{t,i,c}$) as follows:
\begin{equation}
\begin{aligned}    
L_{ce}^{s}&= -\overset{H\times W}{\underset{i=1}{\sum}}\overset{C}{\underset{c=1}{\sum}} I_{[y^{s}_{i}=c]}log(\hat{y}_{s,i,c})\\
L_{ce}^{t}&= -\overset{H\times W}{\underset{i=1}{\sum}}\overset{C}{\underset{c=1}{\sum}} w_{t,i,c}I_{[\hat{y}^{t}_{i}=c]}log(\hat{y}_{t,i,c})
\label{eq:20}
\end{aligned}
\end{equation}

To reduce pseudo-label noise, we applied confidence weights $w_{t,i,c}=\frac{\overset{H\times W}{\underset{i=1}{\sum}}1_{[\underset{c}{max}\:\:pred^{t}_{i,c}]>\beta}}{H\times W}$ \cite{xie2023sepico}. We adopt the teacher-student architecture and the weights of the teacher network are updated using exponential moving average (EMA) \cite{tarvainen2017mean}.

\subsection{ProtoGMM model}
Our primary goal is to enhance the performance of domain adaptation techniques by improving the alignment of features between source and target domains. While most self-training techniques rely heavily on domain alignment methods, they often neglect the importance of precise domain alignment \cite{xie2023sepico,wang2022cluster}. To tackle this issue, we propose a novel approach called multi-prototype-guided alignments in the embedding space. Our method involves identifying the most representative prototypes per category and utilizing them to align the source and target domains. However, the challenge lies in finding prototypes that can effectively capture the diversity in semantic concepts for each category. If sub-class labels are available, we can use them to define these prototypes for each class. Another possible approach is to utilize global category prototypes from the source domain to guide the alignment between the source and target domains. However, this method has limitations, as global prototypes only capture the common characteristics of each category and do not fully leverage the potential strength of semantic information \cite{xie2023sepico}. Moreover, this approach is based on the unimodality assumption of each category, ignoring within-class variations. To overcome these challenges, we introduce the ProtoGMM approach, which aims to address the issues associated with existing methods and improve domain alignment for enhanced domain adaptation. In this approach, we estimate the underlying multi-prototype source distribution by employing the GMM model on the feature space of the source samples. The GMM components serve as the most suitable representative prototypes. The GMM model adapts effectively to multimodal data density, capturing within-class variations. In this approach, to increase intra-class semantic similarity and decrease inter-class similarity across the source domain, we compute the multi-prototype CL loss between source pixel embeddings and the source prototypes. As illustrated in Algorithm \ref{alg:algorithm}, during each iteration, we first update the Gaussian Mixture Model (GMM) of the source pixel data distribution using the high-dimensional $l_2$-normalized features $f_s$ of the source pixels. Subsequently, we compute the multi-prototype CL for the source sample features ($f_{s}$). For the semantic alignment of the source and target domain on the feature space, we propose to perform CL between the embedding of the target samples and source domain multi-prototypes, as shown in Algorithm \ref{alg:algorithm}. Since the model is biased toward the source domain and exhibits a discrepancy between the source and target domains, we propose an alignment mechanism to identify reliable prototypes for the given target sample. The proposed approach's details are elaborated in the following sections.

\begin{algorithm}[h]
\caption{ProtoGMM model}  
\label{alg:algorithm}
\textbf{Input}: Input source instances $(X_{s})$, and associated Labels $(Y_{s})$, Input target instances $(X_{t})$
\begin{algorithmic}[1] 
\STATE Initialize the weights of the model.
\FOR{Iter = 1: $N_{Iter}$}
\FOR{$n\in{1,...,N_{batch}^{s}}$(source minibatch)}
\STATE Update source pixel data distribution GMM model ${\{\phi^{*}_{c}\}}$ using Sinkhorn EM
\IF{$Iter>Iter_{dist}$}
\STATE Apply source domain multi-prototype CL for $f_{s}$
\ENDIF
\STATE Update the source prior distribution $\delta_{source}^{c}$
\ENDFOR
\FOR{$n\in{1,...,N_{batch}^{t}}$(target minibatch)}
\STATE Update the target prior distribution $\delta_{target}^{c}$
\STATE Update target bank by choosing reliable $f_{t}$
\IF{$Iter>Iter_{dist}$}
\STATE Aligning source and target domain by applying target domain multi-prototype CL for $f_{t}$
\ENDIF
\ENDFOR
\ENDFOR
\end{algorithmic}
\end{algorithm}

\subsection{Multiprototype source domain distribution}

The goal is to represent the Multiprototype source domain joint distribution $p(f_{s}, c)$ in the latent feature space as shown in Figure \ref{fig:approach}-a. To achieve this, we can approximate the joint distribution by estimating the class conditional distribution $p(f_{s}|c)$ together with the class prior $p(c)$:
\begin{equation}
\begin{aligned}    
p(f_{s},c)=p(f_{s}|c)p(c)
\label{eq:1}
\end{aligned}
\end{equation}
In our approach, we establish uniform class source priors, achieved through the adoption of a class-balanced sampling technique called rare class sampling (RCS) proposed in \cite{hoyer2022daformer}. By employing RCS, each class is equally represented during training, leading to a balanced distribution. To further enhance the capabilities of our model, we employ generative GMMs to estimate the class-conditional distribution $p(f_{s}|c)$ for each category. This innovative technique enables ProtoGMM to adapt adeptly to datasets with multiple modes of data densities. The GMM consists of a weighted mixture of M multivariate Gaussians, which effectively models the pixel data distribution of each class c in the D-dimensional feature space, as shown in Equation \ref{eq:2}. 
\begin{equation}
\begin{aligned}    p(\boldsymbol{f_{s}}|c;\boldsymbol{\phi_{c}})&=\Sigma_{m=1}^{M}p(m|c;\boldsymbol{\pi}_{c})p(\boldsymbol{f_{s}}|c,m;\boldsymbol{\mu_{c}},\boldsymbol{\Sigma_{c}})\\
    &=\Sigma_{m=1}^{M}\pi_{cm}\mathcal{N}(\boldsymbol{f_{s}};\boldsymbol{\mu_{cm}},\boldsymbol{\Sigma_{cm}}) 
\label{eq:2}
\end{aligned}
\end{equation}

where, $\pi_{cm}$ is a prior probability for each class and $\Sigma_{m=1}^{M}\pi_{cm}=1$; $\Sigma_{c}$ and $\mu_{c}$ are the covariance matrix and mean vector. The GMM classifier is parameterized by $\{\phi_{c}^{*}=\{\boldsymbol{\pi_{c}}, \boldsymbol{\mu_{c}}, \boldsymbol{\Sigma{c}}\}\}_{c=1}^{C}$ and is optimized online using a momentum-based version of the (Sinkhorn) EM (Expectation-Maximization) algorithm, as proposed in \cite{liang2022gmmseg}. The objective of the EM method is to maximize the log-likelihood over the feature-label pairs, which is expressed as ${\phi^{*}_{c}}=\underset{\phi_{c}}{\arg\max}\underset{{f_{s}:y_{s}=c}}{\sum}log{\sum^{M}_{m=1}} p(f_{s},m|c;\phi_{c})
$

\subsection{Source domain multi-prototype CL}

We apply CL between source pixel embeddings and class prototypes, calculated as the means of GMM components for each class, using Equation \ref{eq:4}. The question that arises is how to select negative and positive prototypes for the given source sample.
As previous works have confirmed the significance of hard negatives in metric learning \cite{kaya2019deep}, we propose a hard sampling mechanism to enable multi-prototype CL on the source domain. Employing Bayes' rule and assuming uniform class priors, we compute the posterior using Equation \ref{eq:5}. $p_{s}(m|f_{s},c;\phi^{*}_{c})$ indicates the likelihood of data $f_{s}$ being assigned to component $m$ in class $c$.

Based on the values of $p_{s}(m|f_{s},c;\phi^{*}_{c})$ for the given source sample, considering its ground truth label $c$, we choose the prototype corresponding to the mean of the closest component with the same label as a positive prototype (Equation \ref{eq:6}). Moreover, from the GMM distribution of the categories with different labels, we select the closest component as the hardest component for each category as hard negative prototypes:
\begin{equation}
\begin{aligned}    
l_{protoGMM} = -log \frac{e^{f_{s}q^{+}/\tau}}{e^{f_{s}q^{+}/\tau}+\Sigma_{n=1}^{N}e^{f_{s}q^{-}_{c}/\tau}}
\label{eq:4}
\end{aligned}
\end{equation}
\begin{equation}
\begin{aligned}    
 p_{s}(m|f_{s},c;\phi^{*}_{c})=\frac{\pi_{c,m}\mathcal{N}(f_{s}|\mu_{c,m},\Sigma_{c,m})}{\Sigma_{m'=1}^{M}\pi_{c,m'}\mathcal{N}(f_{s}|\mu_{c,m'},\Sigma_{c,m'})}
\label{eq:5}
\end{aligned}
\end{equation}
\begin{equation}
\begin{aligned}    
q^{+}&=\{\mu_{c,m^{+}}|\:\:m^{+}=\underset{m}{\arg\max}\:\:p_{s}(m|f_{s},c;\phi^{*}_{c}), c=y^{s}\}
\label{eq:6}
\end{aligned}
\end{equation}
\begin{equation}
\begin{aligned}    
q^{-}_{c}&=\{\mu_{c,m^{-}}|\:\:m^{-}=\underset{m}{\arg\max}\:\:p_{s}(m|f_{s},c;\phi^{*}_{c}),c\} \forall c \neq y^{s}
\label{eq:7}
\end{aligned}
\end{equation}

\subsection{Prior distribution update}
We update the prior distribution of the target and source domain using Equation \ref{eq:8}. The equation functions for both domains and the index $d$ indicates whether it pertains to the source or target domains. Where, $\delta_{Iter}^{c}$ denotes the proportion of pixels belonging to the c-th category in the given iteration,$N_{batch}^{d}$ is a number of the images in the given minibatch, $H \times W$ is the multiplication of the height and width of the image indicating the total number of the pixels, $y^{d}_{n, i}$ shows the pixel's ground truth label for the source domain and pseudo label for the target domain. 
\begin{equation}
\begin{aligned}    
\delta_{Iter}^{c}&=\frac{1}{N_{batch}^{d}\times H \times W}\Sigma_{n=1}^{N_{batch}^{d}} \Sigma_{i=1}^{H \times W} y^{d}_{n,i}\\
\delta_{d}^{c}&=\alpha\delta_{d}^{c}+(1-\alpha)\delta_{Iter}^{c}\:\:\:\:\:\:
\:\:\:\:\:\:\:\:\:\:\:\:\forall d=\{s,t\}
\label{eq:8}
\end{aligned}
\end{equation}

\subsection{Update target bank}
The target bank is updated in each iteration by incorporating reliable pixel embeddings from each target mini-batch. To identify these reliable embeddings, we begin by computing the average pixel embedding per class within the given mini-batch, utilizing their pseudo labels (Equation \ref{eq:9}). Next, we assess their cosine similarity with the average mean per class and select the M pixel embeddings with the highest cosine similarity scores as the most reliable representations. 
\begin{equation}
\begin{aligned}    
\mu'^{c}_{t}&=\frac{\Sigma_{i}^{N_{batch}^{t}\times H \times W} f_{t,i}\times I(\hat{y}_{t,i}=c)}{\Sigma_{i}^{N_{batch}^{t}\times H \times W} I(\hat{y}_{t,i}=c)}
\label{eq:9}
\end{aligned}
\end{equation}

\subsection{Target domain prototypes}

We estimate the underlying distribution of the target domain by computing the class prototypes, as shown in Figure \ref{fig:approach}-b. The target domain prototypes will be updated using the target memory bank and EMA: $\mu_{t}^{c}=\alpha\mu_{t}^{c}+(1-\alpha)\mu'^{c}_{t}$.
Noted, We employ Class-balanced Cropping (CBC) \cite{xie2023sepico} on the unlabeled target image. The CBC encourages the model to prioritize cropping from regions with multiple classes.

\begin{figure}
  \centering 
  \includegraphics[width=4.1in]{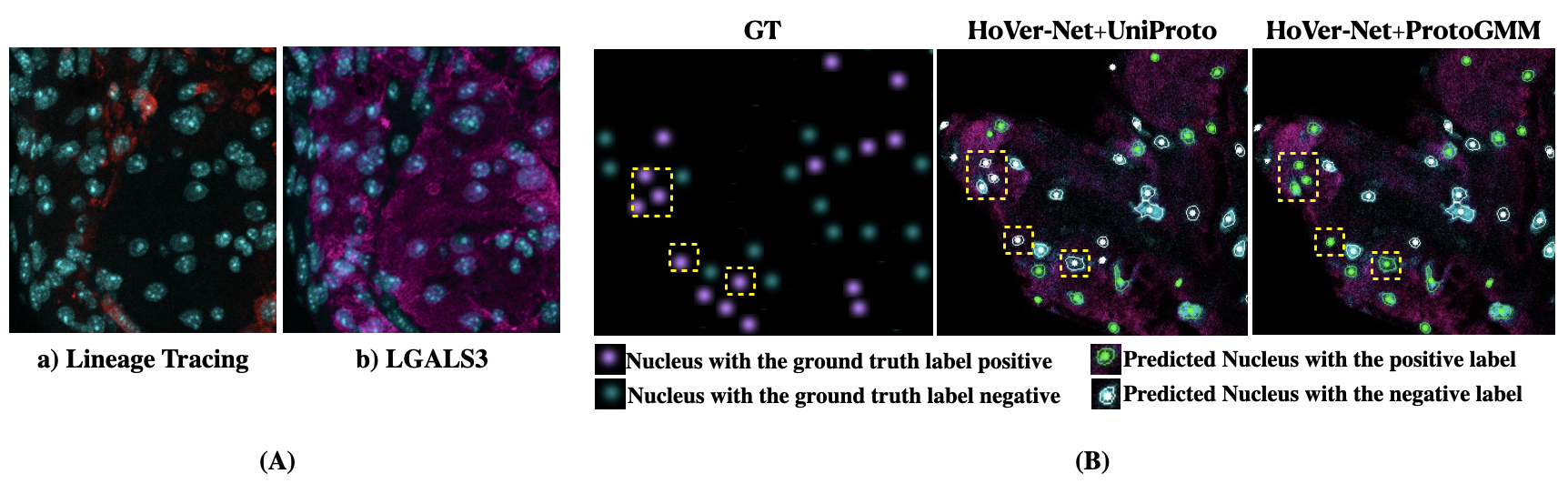}
  \caption{Blue-colored nuclei accompanied by: a) the red Lineage tracing marker, b) the purple LGALS3 marker}
\label{fig:lineage_lgals} 
\end{figure} 
\begin{figure}
  \centering 
  \includegraphics[width=4in]{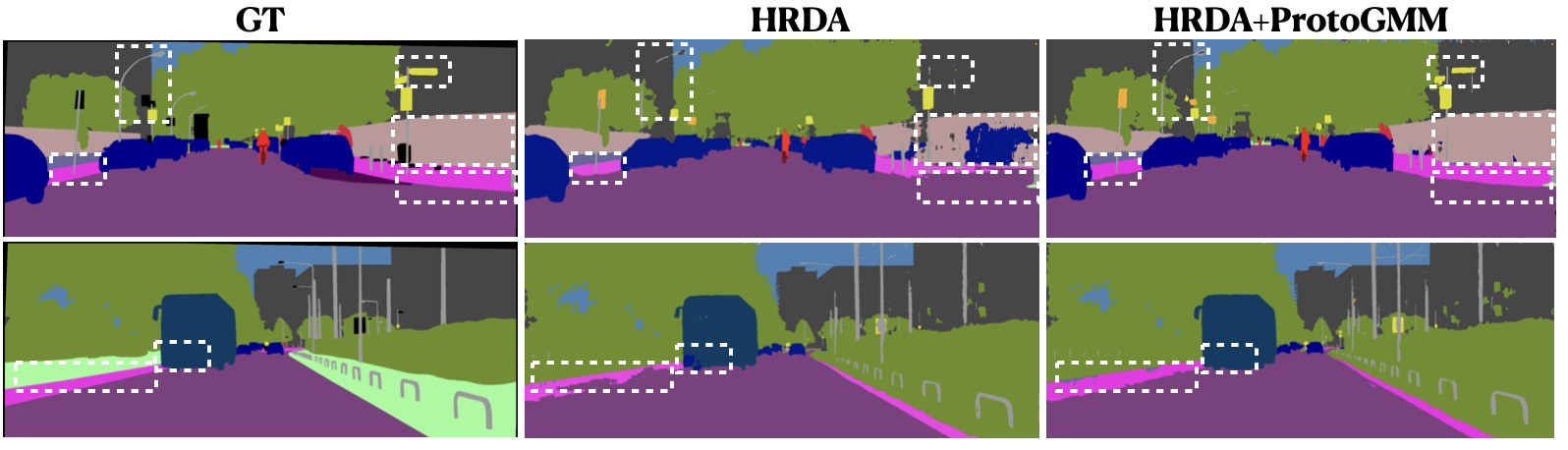}
  \caption{Qualitative analysis on GTA $\rightarrow$ Cityscapes (first row) and Synthia $\rightarrow$ Cityscapes (second row).}
\label{fig:comparison} 
\end{figure} 
\subsection{Aligning source and target domain distribution}

To align the distributions of the source and target domains, we employ multi-prototype CL between the target pixel embeddings and source multi-prototypes (Figure \ref{fig:approach}-c). Since the true labels of target samples are unavailable, we assign a pseudo label to the given target sample using the posterior probability $p_{t}(c|f_{t};\phi^{*}_{c})$ and its similarity to the target prototypes, as shown in Equation \ref{eq:11}. 
For instance, the orange-colored sample is near the target prototype of class 1 and the source component 3 of class 1, as illustrated in Figure \ref{fig:approach}-c. 
The posterior probability indicates the likelihood of the given target sample belonging to class c. The justification for using the posterior probability to assign pseudo labels lies in the domain closeness assumption. This assumption suggests that features from two domains are clustered in a shared space, wherein clusters with identical semantic labels are situated close to each other \cite{wang2022cluster}. Given this premise, we suggest that the target sample should be close to the source domain distribution of the same category within the feature space.
\begin{equation}
\begin{aligned}  
\hat{y}_{t}=\underset{c}{\arg\max}\:\:\:p_{t}(c|f_{t};\phi^{*}_{c})\times \frac{e^{cosine(\mu^{t}_{c},f_{t})}}{\Sigma_{c'}e^{cosine(\mu^{t}_{c'},f_{t})}}
\label{eq:11}
\end{aligned}
\end{equation}

where the first term is the posterior probability and is computed using Proposition 1; the second term computes the cosine similarity of the given target sample with the target prototype. 
The second term corrects the posterior probabilities of class c for the given target sample based on its similarity to the target prototype of class c.

\textbf{Proposition 1:} Given $p_{s}(c|f_{t};\phi^{*}_{c})=\underset{m'}{\sum}{p_{s}(c,m'|f_{t};\phi^{*}_{c})}$, the posterior probability for the given target sample $f_{t}$ is computed as  follows:

\begin{equation}
\begin{aligned}  
p_{t}(c|f_{t};\phi^{*}_{c})&=p_{s}(c|f_{t};\phi^{*}_{c})\times \frac{\delta_{target}^{c}}{\delta_{source}^{c}}
\label{eq:12}
\end{aligned}
\end{equation}
Noted, adjusting the posterior using the ratio $\frac{\delta_{target}^{c}}{\delta_{source}^{c}}$ addresses the issue of label shift as noted in \cite{liu2022undoing}.

\begin{proof}
Based on the Bayes rule, We have the below relationship for the source and target posterior probabilities: 
\begin{equation}
\begin{aligned}  
p_{t}(c|f_{t};\phi^{*}_{c})\:&\alpha \:p_{t}(f(x)|c;\phi^{*}_{c})p_{t}(c)\\
p_{s}(c|f_{t};\phi^{*}_{c})\:&\alpha \:p_{s}(f(x)|c;\phi^{*}_{c})p_{s}(c)
\label{eq:13}
\end{aligned}
\end{equation}

If we assume the conditional data distribution is well aligned, i.e. $p_{t}(f(x)|c)=p_{s}(f(x)|c)$. We can extract the below relationship between the posterior probabilities:  
\begin{equation}
\begin{aligned}  
p_{t}(c|f_{t};\phi^{*}_{c})\:=
p_{s}(c|f_{t};\phi^{*}_{c})\times \frac{p_{t}(c)}{p_{s}(c)}\:
=p_{s}(c|f_{t};\phi^{*}_{c})\times \frac{\delta_{target}^{c}}{\delta_{source}^{c}}
\label{eq:14}
\end{aligned}
\end{equation}
\end{proof}
To perform the multi-prototype CL between the target pixel embeddings and source multi-prototypes, the positive/negative prototypes are chosen as:  
\begin{equation}
\begin{aligned}  
q^{+}=\{\mu_{c,m^{+}}|\:\:m^{+}=\underset{m}{\arg\max}\:\:p_{t}(m|f_{t},c;\phi^{*}_{c})
, c=\hat{y}_{t}\}\\
q^{-}_{c}=\{\mu_{c,m^{-}}|\:\:m^{-}=\underset{m}{\arg\max}\:\:p_{t}(m|f_{t},c;\phi^{*}_{c})
,c\} \forall c \neq \hat{y}_{t}
\label{eq:15}
\end{aligned}
\end{equation}
\section{Experiments}

\subsection{Datasets}

\begin{itemize}
    \item \textbf{Cityscapes \cite{cordts2016cityscapes}:} This dataset comprises real urban scenes captured across 50 cities in Germany and nearby nations. This dataset has segmentation masks with 19 distinct categories for 2,975 training images and 500 validation images, all at a resolution of 2048×1024 pixels. We use the unlabeled training set as a target domain in our experiments, while evaluations are conducted using the validation set.
    
    \item \textbf{GTA \cite{richter2016playing}:} The dataset comprises 24,966 synthetic images extracted from the open-world game "Grand Theft Auto V", each accompanied by corresponding semantic segmentation maps. These images have a resolution of 1914x1052. This dataset is used as a source domain and has 19 common classes with the Cityscapes.

    \item \textbf{Synthia \cite{ros2016synthia}:} It is a synthetic dataset encompassing 9400 photo-realistic frames with a resolution of 1280 × 960. These frames, rendered from a virtual, are paired with pixel-level annotations. This dataset serves as the source domain and shares 13 common semantic annotations with the Cityscapes dataset.

    \item \textbf{Cardiovascular dataset:} The dataset consists of 26 3D multi-channel immunofluorescent images showcasing advanced atherosclerotic lesions from two mouse models. For training and testing, 13 images are allocated to each set. These images encompass multiple channels, including those for nuclei and various markers such as Lineage Tracing, LGALS3, etc., as illustrated in Figure \ref{fig:lineage_lgals}-A. The analysis of these images facilitates the identification of different cell types based on the overlapping presence of nuclei and various markers. The labeling process assigns a "positive" label only when the detected nucleus precisely aligns with the designated marker, as detailed in \cite{moradinasab2023label}. In our evaluation, we designate the Lineage Tracing marker as the source domain and LGALS3 as the target domain, utilizing a subset of 13 images for assessment.
\end{itemize}
\begin{table}
\caption{Comparison with state-of-the-art methods for UDA}
\centering
\scalebox{0.8}{\begin{tabular}{|p{1.8cm}|*{20}{@{\hspace{0pt}}c@{\hspace{0pt}}|}}
\hline
\multicolumn{21}{|c|}{\textbf{GTA5}$\rightarrow$\textbf{Cityscapes}}\\
\hline
Model&	Road&	S.W.&	Build.&	Wall&Fence&	Pole&	T.Li.&T.Sign&	Veget.&	Ter.&Sky&	Person&	Rider&	Car&	Truck&	Bus&Train&	M.Bike&	Bike&	mIoU\\
\hline
AdaptSeg&	86.5&	25.9&79.8&22.1&20.0&23.6&	33.1&21.8&81.8&	25.9&75.9&57.3&26.2&76.3&29.8&	32.1&7.2&29.5&32.5&	41.4\\
\hline
DACS &89.9&39.7&87.9&30.7&39.5&38.5&46.4&52.8&88.0&44.0&88.8&67.2&35.8&84.5&45.7&50.2&0.0&27.3&34.0&52.1\\
\hline
BAPA&94.4&61.0&88.0&26.8&39.9&38.3&46.1&55.3&87.8&46.1&89.4&68.8&40.0&90.2&60.4&59.0&0.0&45.1&54.2&57.4\\
\hline
ProDA&87.8&56.0&79.7&46.3&44.8&45.6&53.5&53.5&88.6&45.2&82.1&70.7&39.2&88.8&45.5&59.4&1.0&48.9&56.4&57.5\\
\hline
DAFormer&95.7&70.2&89.4&53.5&48.1&49.6&55.8&59.4&89.9&47.9&92.5&72.2&44.7&92.3&74.5&78.2&65.1&55.9&61.8&68.3\\
\hline
\SetCell{}{DAFormer+ \textbf{ProtoGMM}}& 
\multirow{2}{*}{\textbf{97.3}}&\multirow{2}{*}{\textbf{79.5}}&\multirow{2}{*}{90.1}&\multirow{2}{*}{55.6}&\multirow{2}{*}{52.2}&\multirow{2}{*}{53.7}&\multirow{2}{*}{58.18}&\multirow{2}{*}{63.4}&\multirow{2}{*}{90.53}&\multirow{2}{*}{49.5}&\multirow{2}{*}{91.83}&\multirow{2}{*}{74.6}&\multirow{2}{*}{46.42}&\multirow{2}{*}{93.3}&\multirow{2}{*}{73.21}&\multirow{2}{*}{80.0}&\multirow{2}{*}{68.74}&\multirow{2}{*}{53.8}&\multirow{2}{*}{65.41}&\multirow{2}{*}{70.4}\\
\hline
HRDA&96.4&74.4&91&\textbf{61.6}&51.5&57.1&\textbf{63.9}&69.3&\textbf{91.3}&48.4&94.2&79.0&52.9&93.9&84.1&85.7&75.9&63.9&\textbf{67.5}&73.8\\
\hline
\SetCell{}{HRDA+ \textbf{ProtoGMM}}& 
\multirow{2}{*}{96.83}&\multirow{2}{*}{77.3}&\multirow{2}{*}{\textbf{90.9}}&\multirow{2}{*}{60.5}&\multirow{2}{*}{\textbf{55.1}}&\multirow{2}{*}{\textbf{59.9}}&\multirow{2}{*}{62.9}&\multirow{2}{*}{\textbf{73.59}}&\multirow{2}{*}{90.8}&\multirow{2}{*}{\textbf{50.0}}&\multirow{2}{*}{\textbf{94.8}}&\multirow{2}{*}{\textbf{79.2}}&\multirow{2}{*}{\textbf{53.5}}&\multirow{2}{*}{\textbf{94.7}}&\multirow{2}{*}{\textbf{86.8}}&\multirow{2}{*}{\textbf{89.5}}&\multirow{2}{*}{\textbf{78.2}}&\multirow{2}{*}{\textbf{65.3}}&\multirow{2}{*}{67.3}&\multirow{2}{*}{\textbf{75.1}}\\
\hline
\multicolumn{21}{|c|}{\textbf{Synthia}$\rightarrow$\textbf{Cityscapes}}\\
\hline
Model&	Road&	S.W.&	Build.&	Wall&Fence&	Pole&	T.Li.&T.Sign&	Veget.&	Ter.&Sky&	Person&	Rider&	Car&	Truck&	Bus&Train&	M.Bike&	Bike&	mIoU\\
\hline
AdaptSeg&	79.2&37.2&78.8&-&-&-&9.9&10.5&78.2&-&80.5&53.5&19.6&67.0&-&29.5&-&21.6&31.3&37.2\\
\hline
DACS&	80.6&25.1&81.9&21.5&2.9&37.2&22.7&24.0&83.7&-&90.8&67.5&38.3&82.9&-&38.9&-&28.5&47.6&48.3\\
\hline
BAPA&	91.7&53.8&83.9&22.4&0.8&34.9&30.5&42.8&86.6&-&88.2&66.0&34.1&86.8&-&51.3&-&29.4&50.5&53.3\\
\hline
ProDA&93.3&61.6&85.3&19.6&5.1&37.8&36.6&42.8&84.9&-&90.4&69.7&41.8&85.6&-&38.4&-&32.6&53.9&55.0\\
\hline
DAFormer&84.5&40.7&88.4&41.5&6.5&50.0&55.0&54.6&86.0&-&89.8&73.2&48.2&87.2&-&53.2&-&53.9&61.7&60.9\\
\hline
\SetCell{}{DAFormer+ \textbf{ProtoGMM}}& 
\multirow{2}{*}{\textbf{93.4}}&\multirow{2}{*}{\textbf{64.3}}&\multirow{2}{*}{87.8}&\multirow{2}{*}{23.5}&\multirow{2}{*}{\textbf{14.1}}&\multirow{2}{*}{53.6}&\multirow{2}{*}{60.1}&\multirow{2}{*}{59.4}&\multirow{2}{*}{86.3}&\multirow{2}{*}{-}&\multirow{2}{*}{88.6}&\multirow{2}{*}{65.2}&\multirow{2}{*}{49.5}&\multirow{2}{*}{89.3}&\multirow{2}{*}{-}&\multirow{2}{*}{62.3}&\multirow{2}{*}{-}&\multirow{2}{*}{52.3}&\multirow{2}{*}{63.6}&\multirow{2}{*}{63.3}\\
\hline
HRDA&85.2&47.7&\textbf{88.8}&\textbf{49.5}&4.8&57.2&65.7&60.9&85.3&-&92.9&79.4&52.8&89.0&-&\textbf{64.7}&-&63.9&64.9&65.8\\
\hline
\SetCell{}{HRDA+ \textbf{ProtoGMM}}& 
\multirow{2}{*}{91.9}&\multirow{2}{*}{59.2}&\multirow{2}{*}{88.7}&\multirow{2}{*}{46.2}&\multirow{2}{*}{4.5}&\multirow{2}{*}{\textbf{59.6}}&\multirow{2}{*}{\textbf{66.6}}&\multirow{2}{*}{\textbf{62.3}}&\multirow{2}{*}{\textbf{87.5}}&\multirow{2}{*}{-}&\multirow{2}{*}{\textbf{94.1}}&\multirow{2}{*}{\textbf{81.1}}&\multirow{2}{*}{\textbf{57.8}}&\multirow{2}{*}{\textbf{91.5}}&\multirow{2}{*}{-}&\multirow{2}{*}{50.4}&\multirow{2}{*}{-}&\multirow{2}{*}{\textbf{65.3}}&\multirow{2}{*}{\textbf{66.9}}&\multirow{2}{*}{\textbf{67.1}}\\
\hline
\end{tabular}}
\label{table:resulttable}
\end{table}

\subsection{Implementation Details}
\textbf{Network architecture:} 
For DAFormer+ProtoGMM and HRDA+ProtoGMM, we adopt the same framework and mainstream pipelines as suggested in  \cite{hoyer2022daformer} and \cite{hoyer2022hrda}, respectively.
Furthermore, we incorporate two $1 \times 1$ convolutional layers with ReLU \cite{xie2023sepico} as a projection head into the network, which maps the high-dimensional pixel embeddings into a 64-dimensional $l-2$ normalized feature vector (D=64). For HoVer-Net+ProtoGMM, we adopt the same framework and settings as suggested in \cite{moradinasab2023label}, with D set to 256. Also, the covariance matrices $\Sigma \in R^{D \times D}$ used in GMM are restricted to a diagonal structure.

\noindent
\textbf{Training:} 
We follow the same training regime as described in \cite{hoyer2022daformer} and \cite{hoyer2022hrda}, and \cite{moradinasab2023label} for DAFormer+ProtoGMM, HRDA+ProtoGMM, and HoVer-Net+ProtoGMM, respectively.
All models are developed using PyTorch 1.8.1 and trained on a single NVIDIA Tesla V100-32G GPU. 
In DAFormer+ProtoGMM and HRDA+ProtoGMM, We used the AdamW optimizer \cite{loshchilov2017decoupled} and set the betas set and weight decay at (0.9, 0.999) and 0.01. We incorporate the learning rate warmup policy as same as \cite{hoyer2022daformer}. We set $\alpha$ to 0.9 and the EMA weight update parameter ($\beta$) for the teacher network to 0.999. We train the the DAFormer+ProtoGMM model for 60K epochs and the HRDA+ProtoGMM model for 80K epochs. In HoVer-Net+ProtoGMM, we used the Adam optimizer \cite{loshchilov2017decoupled} with the learning rate 1.0e-4 and set the betas at (0.9, 0.999). For the generative optimization of GMM, we follow the same framework from GMMSeg \cite{liang2022gmmseg}. In each iteration, we perform one iteration of the momentum (Sinkhorn) Expectation-Maximization (EM) process, on both the current training batch and the external memory. The size of external memory is 32K pixel features per category. For the DAFormer model, components per category are set to 5 and 3 for GTA and Synthia datasets, respectively. For HoVerNet and HRDA models, these values are 3 and 5. This memory is updated by the first-in, first-out matter and by selecting 100 pixels per class from each image. The size of the target memory bank is 16K per class. Our evaluation metric employs per-class intersection-over-union (IoU), mean IoU, recall, and F1-score. 

\subsection{Comparison with existing UDA methods}
We compare the proposed ProtoGMM with existing methods in two different scenarios: 1) the synthetic-to-real adaptation scenario, and 2) the cell-type adaptation scenario.

\textbf{The synthetic-to-real adaptation scenario:}
We show that the proposed ProtoGMM improves the performance of existing UDA methods in two representative synthetic-to-real adaptation scenarios: GTA5 $\rightarrow$ Cityscapes and Synthia $\rightarrow$ Cityscapes in Table \ref{table:resulttable}. The UDA methods include AdaptSeg \cite{tsai2018learning}, DACS \cite{tranheden2021dacs}, BAPA \cite{liu2021bapa}, ProDA \cite{zhang2021prototypical}, DAFormer \cite{hoyer2022daformer}, and HRDA \cite{hoyer2022hrda}. Our results reveal that DAFormer+ProtoGMM surpasses the current method by a notable margin of +2.1 mIoU in the GTA5 $\rightarrow$ Cityscapes scenario and +2.4 mIoU in the Synthia $\rightarrow$ Cityscapes case, as outlined in Table \ref{table:resulttable}. Additionally, we note that HRDA+ProtoGMM exhibits superior performance over the existing method, achieving a margin of +1.3 mIoU in the GTA5 $\rightarrow$ Cityscapes scenario and +1.29 mIoU in the Synthia $\rightarrow$ Cityscapes case, as detailed in Table \ref{table:resulttable}.
Figure \ref{fig:comparison} shows that the ProtoGMM improves the performance of the HRDA in classes like wall, walk-side, Bus, Sign, etc indicated by white dotted boxes.

\noindent
\textbf{The cell-type adaptation scenario:}
To adapt cell types in immunofluorescent images, we utilized the HoVer-Net+ProtoGMM domain adaptation model, built upon a modified HoVer-Net from \cite{moradinasab2023label}. In our approach, cell types are identified in each image based on their overlap with various markers. A cell is labeled positive for a specific marker if it exhibits overlap with that marker. Treating each cell type as a separate domain, our goal is to enhance the performance of the trained model on one marker for other markers with different distributions, without the need for additional labeling efforts.
In this scenario, the Lineage Tracing marker serves as a labeled source domain, while LGALS3 acts as an unlabeled target domain. We have access to positive/negative labels for nuclei only for the Lineage Tracing marker, while nuclei labels based on LGALS3 are unknown. This adaptation scenario involves both covariate and label shifts, as different markers have distinct distributions (covariate shift), and the distributions of positive and negative cells vary for different markers (label shift).
It's worth noting that we follow the methodology outlined in \cite{moradinasab2023label} to convert these 3D images into 2D within the source domain. In the target domain, we employ a linear combination approach to merge nuclei and marker channels, followed by slicing the images along the z-axis to transform them into 2D images. During model training, we extract patches of size 256×256 pixels with a 10\% overlap from both the source and target domains.
All segmentation pixel-level masks are generated using the point annotation and original images, following the approach introduced in \cite{moradinasab2023label}. Table \ref{table:lineage} highlights that ProtoGMM significantly improves the performance of the base domain adaptation model (HoVer-Net+Self-training), which relies solely on self-training, with increases of +7.31, +5.98, and 5.85 in precision, recall, and F1-score, respectively. Furthermore, HoVer-Net+ProtoGMM outperforms HoVer-Net+UniProto in terms of recall and F1-score while maintaining comparable precision. Additionally, Figure \ref{fig:lineage_lgals}-B presents a qualitative analysis of Lineage Tracing $\rightarrow$ LGALS3, illustrating that the HoVerNet+ProtoGMM model accurately predicts the labels of nuclei marked with dashed yellow color.
\begin{table}
\caption{Lineage Training $\rightarrow$ LGALS3}
\centering
\scalebox{1}{\begin{tabular}{|p{3.6cm}|*{3}{@{\hspace{3pt}}c@{\hspace{3pt}}|}}
\hline
Model&Precision&Recall&F1-score \\
\hline
HoVer-Net+Self-training&65.79&70.42&68.05\\
\hline
HoVer-Net+UniProto&77.4&68.9&72.0\\
\hline
HoVer-Net+ProtoGMM&73.1&76.4&73.9\\
\hline
\end{tabular}}
\label{table:lineage}
\end{table}
\begin{table}
  \caption{Comparison with SOTA Contrastive learning approachess}
  \centering
  \begin{tabular}{*{1}{|@{\hspace{0pt}}c@{\hspace{0pt}}}*{6}{|@{\hspace{0pt}}c@{\hspace{0pt}}}|}
    \hline
    &\multicolumn{3}{c}{\textbf{GTA5}$\rightarrow$\textbf{City.}} & \multicolumn{3}{c|}{\textbf{Synthia}$\rightarrow$\textbf{City.}}\\
    \hline
    Model&BankCL&UniCL&ProtoGMM&BankCL&UniCL&ProtoGMM\\
    \hline
    mIoU&69.12&69.48&\textbf{70.4}&61.46&61.71&\textbf{63.3}\\
    \hline
  \end{tabular}
  \label{table:cl}
\end{table}

\begin{table}
\caption{Comparison with CL baseline methods}
\centering
\scalebox{1}{\begin{tabular}{|p{1.5cm}|*{3}{@{\hspace{3pt}}c@{\hspace{3pt}}|}}
\hline
Model&UniProto&BankCL&ProtoGMM \\
\hline
mIoU&60.0&59.42&\textbf{61.4}\\
\hline
\end{tabular}}
\label{table:cl2}
\end{table}

\noindent
\textbf{Ablation study:}
Comparing ProtoGMM and pixel contrast methods, UniProto and BankCL, is outlined in Table \ref{table:cl}. UniProto uses global class prototypes as both positive and negative samples, resulting in one positive class and $C-1$ negative classes per sample. In contrast, BankCL utilizes multiple positive and negative samples from a memory bank \cite{xie2023sepico}, involving the storage of local category centroids for individual source images. As Table \ref{table:cl} indicates the ProtoGMM outperforms both BankCL and UniProto methods in both GTA5 $\rightarrow$ Cityscapes and Synthia $\rightarrow$ Cityscapes cases. 
Furthermore, to illustrate our method’s effectiveness, we compared the DAFormer+ProtoGMM model with the DeepLab-V2 backbone to baseline CL methods (BankCL, UniProto) for the GTA5 $\rightarrow$ Cityscapes scenario which confirms the effectiveness of DAFormer+ProtoGMM (Table \ref{table:cl2}).


\section{Conclusion}
In this paper, we introduce the protoGMM model which involves estimating the multi-prototype source distribution by using GMM models in the feature space. The GMM components serve as representative prototypes, effectively adapting to the diversity of the data and capturing variations within classes. To enhance intra-class semantic similarity, reduce inter-class similarity, and align the source and target domains, we apply multi-prototype CL between the source distribution and target samples. Experimental results demonstrate the effectiveness of our approach on the UDA benchmarks.

\subsubsection{Acknowledgements} This work was supported by an American Heart Association Predoctoral Fellowship (23PRE1028980), NIH R01 HL155165, NIH R01 156849, R01HL166161, and National Center for Advancing Translational Science of the National Institutes of Health Award UL1 TR003015. 
\newpage
%
%
%
\bibliographystyle{splncs04}
\bibliography{main}

\end{document}